\documentclass[runningheads]{llncs}
\usepackage[T1]{fontenc}
\usepackage[dvipsnames]{xcolor}
\usepackage{xcolor,colortbl}
\usepackage{graphicx}
\begin{document}
%
\title{Detecting Unforeseen Data Properties with Diffusion Autoencoder Embeddings using Spine MRI data}
%
\titlerunning{Diffusion Autoencoder Embeddings for Bias Detection}

\author{Robert Graf\inst{1,2}\and
Florian Hunecke\inst{1,2}\and
Soeren Pohl\inst{1,2} \and 
Matan Atad\inst{1,2} \and 
Hendrik Moeller\inst{1,2} \and 
Sophie Starck\inst{2}\and 
Thomas Kroencke\inst{4}\and 
Stefanie Bette\inst{4} \and
Fabian Bamberg\inst{4} \and
Tobias Pischon\inst{4} \and
Thoralf Niendorf\inst{4} \and
Carsten Schmidt\inst{5} \and
Johannes C. Paetzold\inst{3}\and
Daniel Rueckert\inst{2,3}\and 
Jan S Kirschke\inst{1}}
\authorrunning{R. Graf et al.}

\institute{Department of Diagnostic and Interventional Neuroradiology, Klinikum rechts der Isar, TUM School of Medicine and Health \and
Institut fuer KI und Informatik in der Medizin, Klinikum rechts der Isar, TUM \and
Biomedical Image Analysis Group, Depart. of Computing, Imperial College London\and
German National Cohort (NAKO) Investigators \and
University Medicine of Greifswald 
}
%
%

\maketitle              

\begin{center}
This paper was accepted in the "Workshop on Interpretability of Machine Intelligence in Medical Image Computing" (iMIMIC) at MICCAI 2024
\end{center}
\begin{abstract}
Deep learning has made significant strides in medical imaging, leveraging the use of large datasets to improve diagnostics and prognostics. However, large datasets often come with inherent errors through subject selection and acquisition. In this paper, we investigate the use of Diffusion Autoencoder (DAE) embeddings for uncovering and understanding data characteristics and biases, including biases for protected variables like sex and data abnormalities indicative of unwanted protocol variations. We use sagittal T2-weighted magnetic resonance (MR) images of the neck, chest, and lumbar region from 11186 German National Cohort (NAKO) participants. We compare DAE embeddings with existing generative models like StyleGAN and Variational Autoencoder. Evaluations on a large-scale dataset consisting of sagittal T2-weighted MR images of three spine regions show that DAE embeddings effectively separate protected variables such as sex and age. Furthermore, we used t-SNE visualization to identify unwanted variations in imaging protocols, revealing differences in head positioning. Our embedding can identify samples where a sex predictor will have issues learning the correct sex. Our findings highlight the potential of using advanced embedding techniques like DAEs to detect data quality issues and biases in medical imaging datasets. Identifying such hidden relations can enhance the reliability and fairness of deep learning models in healthcare applications, ultimately improving patient care and outcomes.
\end{abstract}
%
%
%
\section{Introduction}

Historically, deep learning in medicine has relied on large datasets, which have been difficult to obtain.  However, in recent years, significant progress has been made. Notable examples include extensive datasets such as chest radiographs \cite{irvin2019chexpert,johnson2019mimic}, as well as images from comprehensive epidemiological studies like the UK-Biobank \cite{allen2012uk} and the German National Cohort ("NAKO Gesundheitsstudie") \cite{bamberg2015whole}. Nevertheless, the sheer size of these datasets makes it difficult to detect data shifts and biases visually. Biases can occur through various sources, such as deviations from the examination protocol, the subjects themselves, changes in framing, or data processing errors \cite{fabbrizzi2022survey}. Detecting biases in large data sets is a laborious and time-consuming task and would require a large survey to strongly evaluate the data \cite{hu2019discovering,li2021discover}. For disease prediction, it is paramount to know what biases exist to compensate for them or at least observe them if they have an influence on a classifier. Other approaches tried to reduce biases to force disentangling \cite{berg2022prompt,li2021discover,peebles2020hessian} or use unsupervised embeddings \cite{tommasi2017deeper,elnaggar2021prottrans,petreski2023word}. Unsupervised clustering is a particularly promising option to overcome this hurdle. To better cluster images, we can first reduce the image to an embedding. Then, the whole dataset can be visualized with t-SNE \cite{van2008visualizing}. Embeddings are representations of data generated, in our case, through an unlabeled reconstruction task. In large language models, embeddings are crucial in learning relationships between words, sentence structures, and meaning. For images, these embeddings are often used to manipulate generative networks because they reorder images so that image features, like age or skin tone, are split into individual hyperplanes. We aim to utilize this property of embeddings to gain insight into data properties and thereby identify potential biases. For instance, visualizing embeddings concerning acquisition location or date can reveal biases stemming from differences in acquisition protocols or variations in imaging equipment. Additionally, we can investigate how protected variables like sex and age correlate with the data. Generative embeddings are either generated with the VAEs \cite{higgins2016beta} or StyleGAN2 \cite{karras2020analyzing}. In VAE, the bottleneck output is the embedding of the input. StyleGAN models use inversion networks like "encoder for editing" \cite{tov2021designing} to generate embeddings from real data. The inversion model gets the input and output pairs of StyleGAN and predicts the noise from a given image. Presently, diffusion-based techniques have outperformed GANs, and the Diffusion Autoencoder (DAE) \cite{preechakul2022diffusion} may offer superior embeddings while requiring only a single end-to-end training process. 
\subsubsection{Contributions} We propose DAE \cite{preechakul2022diffusion} embeddings as a better replacement for existing embedding models for medical images. First, we want to investigate if the improved DAE embeddings can better disentangle images than previous generative embeddings. 
Second, we want to highlight the ability to detect data shifts that could impact fairness by investigating two example anomalies we observed in the DAE embeddings and get an explanation for those outliers.

\section{Materials and Methods}
\begin{figure}[htbp]    
  \includegraphics[width=\linewidth]{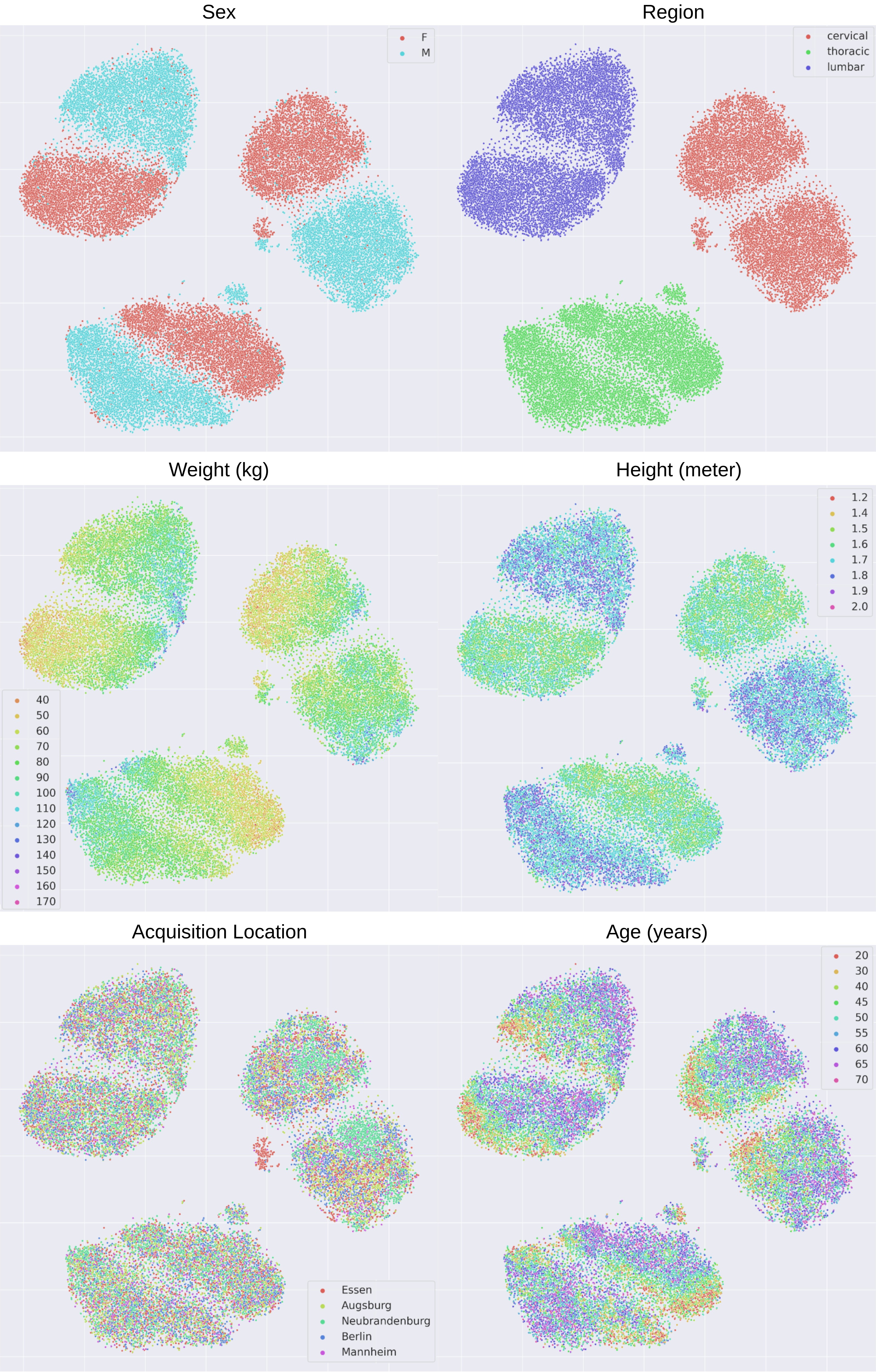}
  \caption{t-SNE plots of our DAE embeddings. The embeddings are colored with the patient sex, height, weight, and age. The Region label describes the body region; the acquisition location is the city where the image was recorded. }\label{tsne}
\end{figure}
This study included 11186 [5745 male; 5441 female] subjects from the NAKO. Each subject provides sagittal T2-weighted (T2w)  magnetic resonance images (MRI) of the neck, chest, and lumbar region. We used sagittal slices to train our DAE. We normalized the data to [-1, 1] for training by linear rescaling from [0, max-value] and random cropping to 256 $\times$ 256. We employed a fixed center cropping approach to create an MRI volume embedding, concatenating 12 center slices into a single embedding. 12 slices are guaranteed to be available, and the number of images must be constant. We want to avoid embedding differences through random cropping; therefore, all embeddings and predictions are made with the same fixed field of view. Three embeddings (neck, chest, lumbar) were evaluated separately for each subject. Subjects were then divided into training, validation, and test sets using an 80\%, 5\%, and 15\% split ratio throughout. The t-SNE visualization encompassed all images. We studied two classes of bias. First, we analyzed individual characteristics like sex, age, weight, and height. These protected variables can cause the prediction to get worse; for instance, in face recognition, most models performed worse for dark-skinned women due to training and testing data imbalance and the difficulty of cameras to collect light from darker skin \cite{najibi2020racial}. The second class of bias is the acquisition location and date. In an ideal scenario, we would be unable to identify the time and location where any of the scans were taken. In a multi-center study that took over three years to collect, there is a risk that we could reidentify different institutes or the time when a scan was taken. We first trained the DAE and extracted the image embeddings. Support Vector Machines (SVMs) \cite{svm} were then trained on balanced embedding data sets. We compared the classification results with other established generative modeling approaches to evaluate the performance of DAE embeddings.

\section{Experiments and Results}

In an initial step, we generated the DAE embeddings and visualized them in a t-SNE plot using a perplexity value of 50 (Figure~\ref{tsne}). We observed that the model can differentiate protected properties even though the process is fully unsupervised. The DAE especially well separates sex, weight, and age. Sex and age are something a radiologist would not be confident in estimating from a spine image alone. The acquisition year had no visible clustering, but the acquisition locations had multiple unexpected clusters in the cervical region. For StyleGAN, we observe the same location-dependent clustering, but the split into sex is not as clear as the DAE embedding in the t-SNE plots.

\subsubsection{Embedding Quality}
To assess embedding quality, we employed SVMs to train a predictor and compare it with networks trained on the images and labels themselves. We used ResNet10, ResNet34, and DenseNet121 to learn the data distribution. We compared DAE with $\beta$-VAE and StyleGAN. To show that DAE outperforms other generative embedding methods, we considered age, weight, height, body region, and sex as training objectives. Body region and sex were treated as a classification task, while age, weight, and height were subject to regression analysis to minimize mean-absolute error. Optimization involved carefully rebalancing the training data to ensure equal representation across all predicted value ranges. This was especially important for age since the networks would otherwise predict the average age for all samples. Further, weight decay between 0.01 to 0.0001 can drastically improve the final prediction of the NN classifiers. Subject age ranges from 20 to 72 years, height ranges from 1.25 to 2.05 meters, and weight varies from  38 to 192 kilograms. All participants self-reported their sex, either male or female. 
\begin{table}[htbp]
\caption{Regression and classification with images and embedding.}
\label{tab:my-table}

\begin{tabular}{lll|cccccc}
                    &\begin{tabular}[c]{@{}c@{}}Super-\\vision\end{tabular} 
                    &Type     
                    &\begin{tabular}[c]{@{}c@{}}Body Region\\ accuracy $\uparrow$\end{tabular} 
                    &\begin{tabular}[c]{@{}c@{}}Sex\\ accuracy $\uparrow$\end{tabular}
                    & \begin{tabular}[c]{@{}c@{}}Weight\\ $\ell_1$ kg $\downarrow$\end{tabular} 
                    & \begin{tabular}[c]{@{}c@{}}Height\\ $\ell_1$ meter $\downarrow$\end{tabular}
                    & \begin{tabular}[c]{@{}c@{}}Age   \\ $\ell_1$ years $\downarrow$\end{tabular}\\ \hline
$\beta$-VAE + Hessian  &semi&embed.&0.998 & 0.870  &  6.75  & 0.055  & 7.80 \\
StyleGAN               &semi&embed.& \textbf{1.000} & 0.885 & 5.66 & 0.047 & 5.62 \\
\rowcolor{GreenYellow}
DAE (ours)             &semi&embed.& \textbf{1.000} & \textbf{0.988} & \textbf{4.32} & \textbf{0.032} &\textbf{ 3.84} \\\hline
ResNet10                &fully&image    & 0.997& 0.993 &10.26 & 0.072 & 4.15 \\
ResNet34                &fully&image    & \textbf{1.000} & \textbf{0.999} & \textbf{3.28} & 0.029 & 3.12 \\
DenseNet121             &fully&image    & 0.997 & \textbf{0.999} &4.09 & \textbf{0.028} & \textbf{2.84} \\ \hline
\end{tabular}
\end{table}
Our DAE embeddings outperformed StyleGAN and $\beta$-VAE by a considerable margin. Only for body regions, StyleGAN also reaches perfect accuracy. The unsupervised embedding and SVM clustering were good enough to beat ResNet10 in all but one task. The larger fully supervised models performed better. Still, the gap between DAE embeddings and supervised classification is smaller than the gap between other embedding methods and DAE embeddings. Our tested age models are in line with other age predictors. For example, dedicated architectures on the full-body excluding the head achieved on the UK-Biobank \cite{allen2012uk} are MAE of 2.38 \cite{he2022deep} and 2.76 \cite{starck2023atlas} with an age range of 44-82. The age range and larger 3D field of view can noticeably impact the MAE, so they are not perfectly comparable. Starck and Kini et al. \cite{starck2023atlas} show that their model mainly focuses on the spine, which is the area we exclusively look at.

\subsection{Bias detection}

In order to investigate the unknown systemic biases, we first delineated clusters in the t-SNE plots. Then, we resorted to training image classifiers on the new cluster labels from the delineation and employed explainability methods to discover the feature's location causing the clustering.

\begin{figure}[htbp]
    
  \includegraphics[width=1\linewidth]{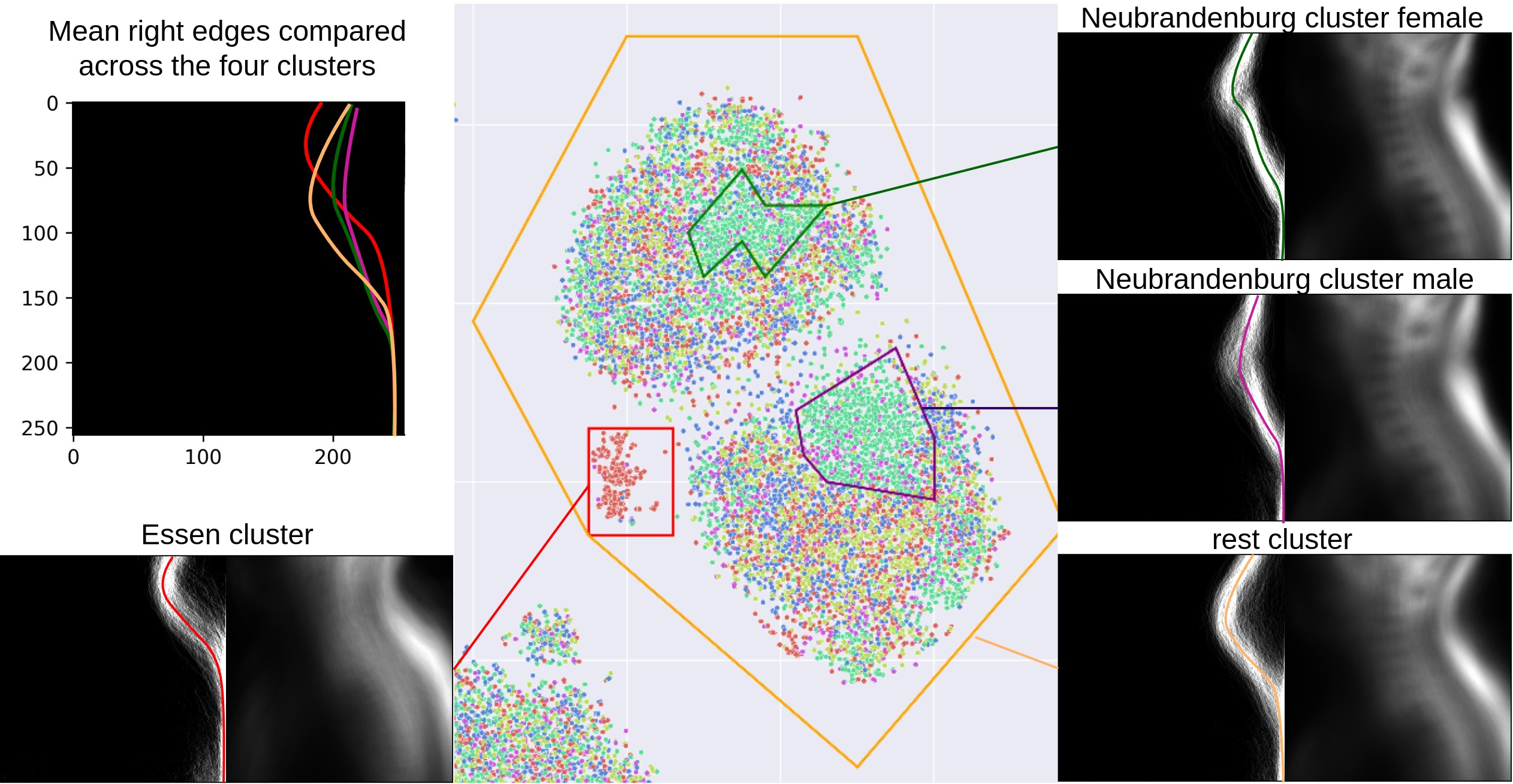}
  \centering
  \caption{Location Biases. Three clusters separate from the head images when we color them by the examination center where the MRI was taken. For each cluster and the rest class, we show two images. The left is a selection of the right edges of the images, and on the right are the images summed together. Both are made from 1000 images and are averaged. We observe that the neck curve differs in the clusters. On the top left, we visualized this by plotting the mean line of all 4 clusters.   \label{fig:headpossition}}
\end{figure}
\subsubsection{Location} To unravel unexpected findings behind unexpected clusters, we implemented a classifier trained on images where we manually delineate cervical location clusters using t-SNE. For better explainability, we used GradCAM\cite{selvaraju2017grad}, which allowed us to visualize the salient features influencing decision-making. The GradCAM activation for the cluster in the cervical area highlighted the right edge (back of the neck). To provide a more straightforward depiction of location bias, we employed an averaging technique across 1000 images from a specific cluster, revealing common shapes and patterns (Figure \ref{fig:headpossition}). We observed that the clusters have other neck curvature by printing the right edge and overlaying 1000 images. We found that the neck is shifted in a constant way in all clusters, which can only be explained by an altered physical or software setup in the scanner. For the red "Essen" cluster, we observed that the whole person is shifted in superior direction by about 50 voxels (43 mm). This could be a technical or human error, where the subjects were sent deeper into the MRI device than other centers or a software issue where the field of view selection diverged from the other scanning locations. The thoracic and lumbar regions, in contrast, are unaffected. The separated green/purple clusters from "Neubrandenburg" and "Mannheim" are created by a difference in a neck rest position determined by the MR device's neck support. The neck curve is much flatter in the "Neubrandenburg" clusters. We also checked for time-dependent clustering and found no clustering. Nonetheless, we observe that subjects in the separated "Neubrandenburg" clusters were scanned in the second and third years but not in the first year. Our embeddings unveiled framing differences in head images. For a subset of images, we can deduce the potential scanning location. In a more extreme case, this would be a privacy concern to be able to re-identify subjects.

\begin{figure}[htbp]
\centering
  \includegraphics[width=1\linewidth]{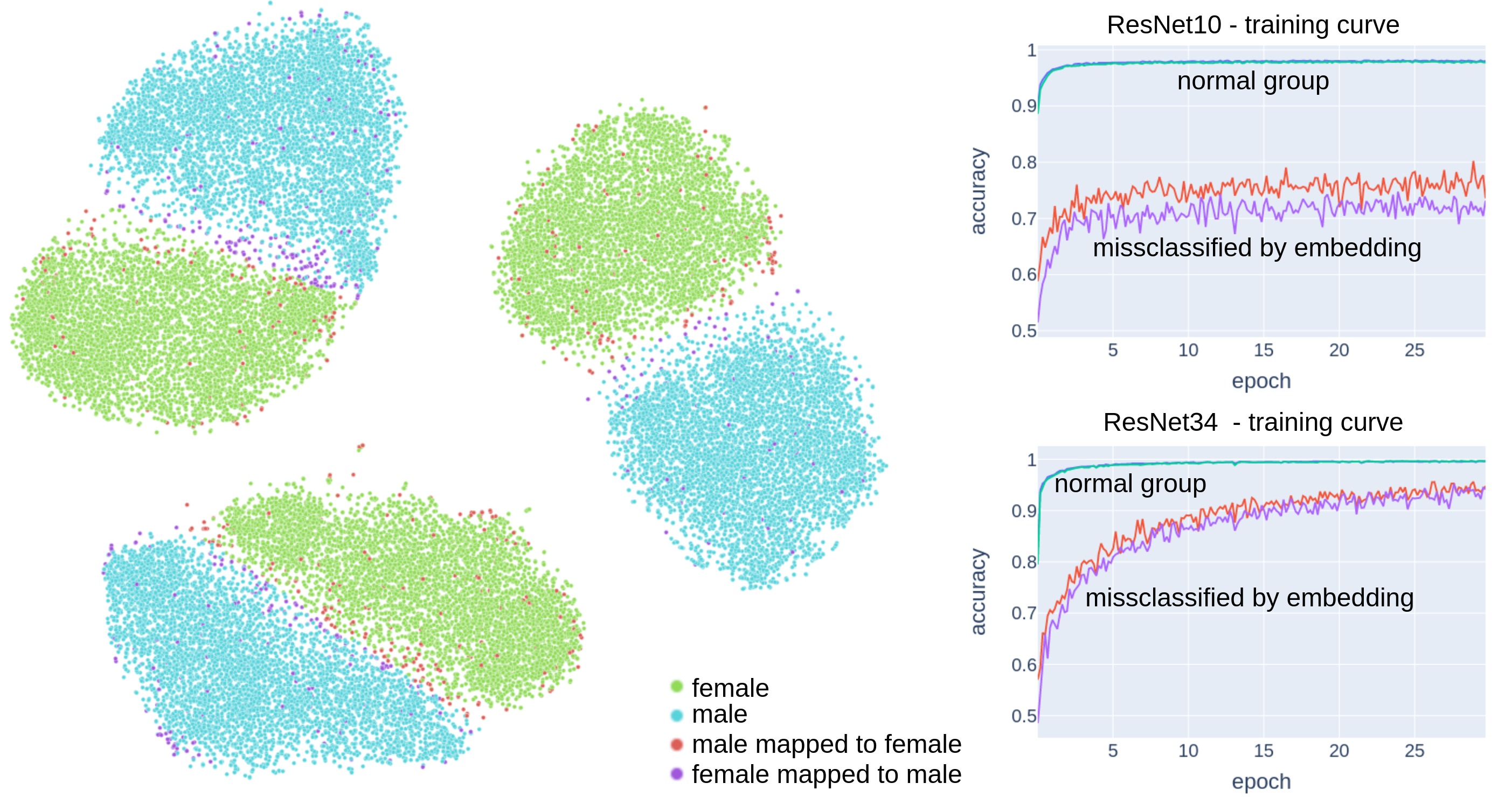}
  \caption{Sex Biases. Left: A subset of subjects are clustered between the male and female blobs, and others are completely pushed to the opposite sex. They must have a set of features that indicate that the spine is the opposite sex.  We have no further evidence of how this is reflected in the sex and gender of those persons overall. Right: The training curves of a male/female classifier of ResNet10 and ResNet34. We train only on the male/female label but measure the misclassified group separately and observe that they clearly lag behind during training.}
  \label{fig:sex}
\end{figure}

\subsubsection{Sex}
The DAE embeddings separate the subjects by sex. We observe that a few hundred people have an embedding that puts their images into the cluster of the other sex. In cases where embeddings deviate from their true sex class, we scrutinized individual instances to detect labeling errors but found that the sexes of the patients were correctly labeled. We defined sex by their primary sex organ as visible in different full-body MRI acquisitions. 411 thoracic (3.7\%), 347 lumbar (3.1\%), and 211 cervical (1.9 \%) images out of 11,186 subjects were classified towards the opposite sex. 
Out of the misclassified images, 585 (77.0 \% of misclassed) had a single region misclassified, 141 (18.6 \%  of misclassed) subjects had two regions that were misclassified, and 34 (4.5 \% of misclassed) subjects had all three regions classified as a different sex. Assuming the independent probability of separate regions, we would expect a count of 26.6 subjects with two misclassified regions and 0.24 subjects for all three regions; therefore, the number of subjects with more than one switch in sex clustering is far above random chance. Misclassifications are equally frequent in both sexes.  
The cervical region probably has the lowest misclassification because, unlike the other images, it has more non-spine tissue visible, especially parts of the skull. The other regions are shadowed towards the front. The back is fully visible in the images. First, we delineated clusters in the t-SNE plot and separated the misclassified images. We trained different networks and observed that the GradCAM highlights the whole spine image. This indicates that it is unclear what image features the network uses because we can not pinpoint a location. We hypothesized that the feature is a texture or structure difference that can be observed in the whole image, thus introducing a potential bias in classification tasks. 
To study this effect, we trained a classifier on the given sex label, but we plot the loss curve for the embedding misclassified group separately. We varied the model and hyperparameter. After one epoch, the classifiers reach 96-98\% and converge to 98\%-99\%, but the misclassified group clearly underperforms. After one epoch, the accuracy is 40-70\% and reaches a maximum of 70-95\%. The validation is consistent with training or swings between overpredicting either male or female. The model picks up on the same feature as the embedding network within the first epoch. Our selected subgroup is clearly biased (See Figure~\ref {fig:sex}) by the classifier and receives worse prediction results than the rest of the population.
Furthermore, we aim to study whether this effect originates from the T2w images alone or from the patient's anatomy. We replaced the T2w image with T1-weighted Dixon technique images. We received the same results with our classifier when using a similar field of view. 
We conclude that we observe a real phenomenon where a network picks up subtle image changes between the two sexes, which are not always reliable for all subjects. The models detect features that can identify the sex of most people, but some persons have the indication of the opposite sex for those features. We believe that this previously unknown bias is a normal variant, and after discussing it with multiple radiologists, we discussed many hormonal and growth differences that could cause this effect but are not visible to the human eye. With large datasets and longer training time, the larger supervised networks can learn additional features to distinguish some of the remaining persons into their correct sex. Still, they take longer to fit into the model, and we expect that smaller datasets do not have enough samples to provide this level of detail.


\section{Conclusion}

We used DAE to visualize properties of T2-weighted MR images of the neck, chest, and lumbar region based on data from the German National Cohort related to individual characteristics and detecting image acquisition abnormalities indicative of protocol violations. We compared DAE with StyleGAN and VAE and showed that DAEs yielded better embeddings. Especially sex, which is dependent on subtle image differences, is better distinguished in DAE embeddings. DAE successfully differentiated sex on spine MRI but were not $100 \%$ identical to the self-reported sex, an issue we can reproduce in a supervised setting, and found a subpopulation that was consistently difficult for a DL-Model to learn. We found inconsistent head positions in the cervical images in different examination centers.  We believe that embedding can be used to find consistent data biases that may hurt learning tasks towards protected properties like sex, age, or race.

\begin{credits}
\subsubsection{\ackname} The research for this article received funding from the European Research Council (ERC) under the European Union’s Horizon 2020 research and innovation program (101045128 — iBack-epic—ERC2021-COG). This project was conducted with data from the German National Cohort (NAKO) (www.nako.de). The NAKO is funded by the Federal Ministry of Education and Research (BMBF) [project funding reference numbers: 01ER1301A/B/C, 01ER1511D and 01ER1801A/B/C/D], federal states of Germany and the Helmholtz Association, the participating universities and the institutes of the Leibniz Association. We thank all participants who took part in the NAKO study and the staff of this research initiative.
\end{credits}
\bibliographystyle{splncs04}
\bibliography{mybibliography}

\newpage
\begin{figure}[htbp]    
  \includegraphics[width=\linewidth]{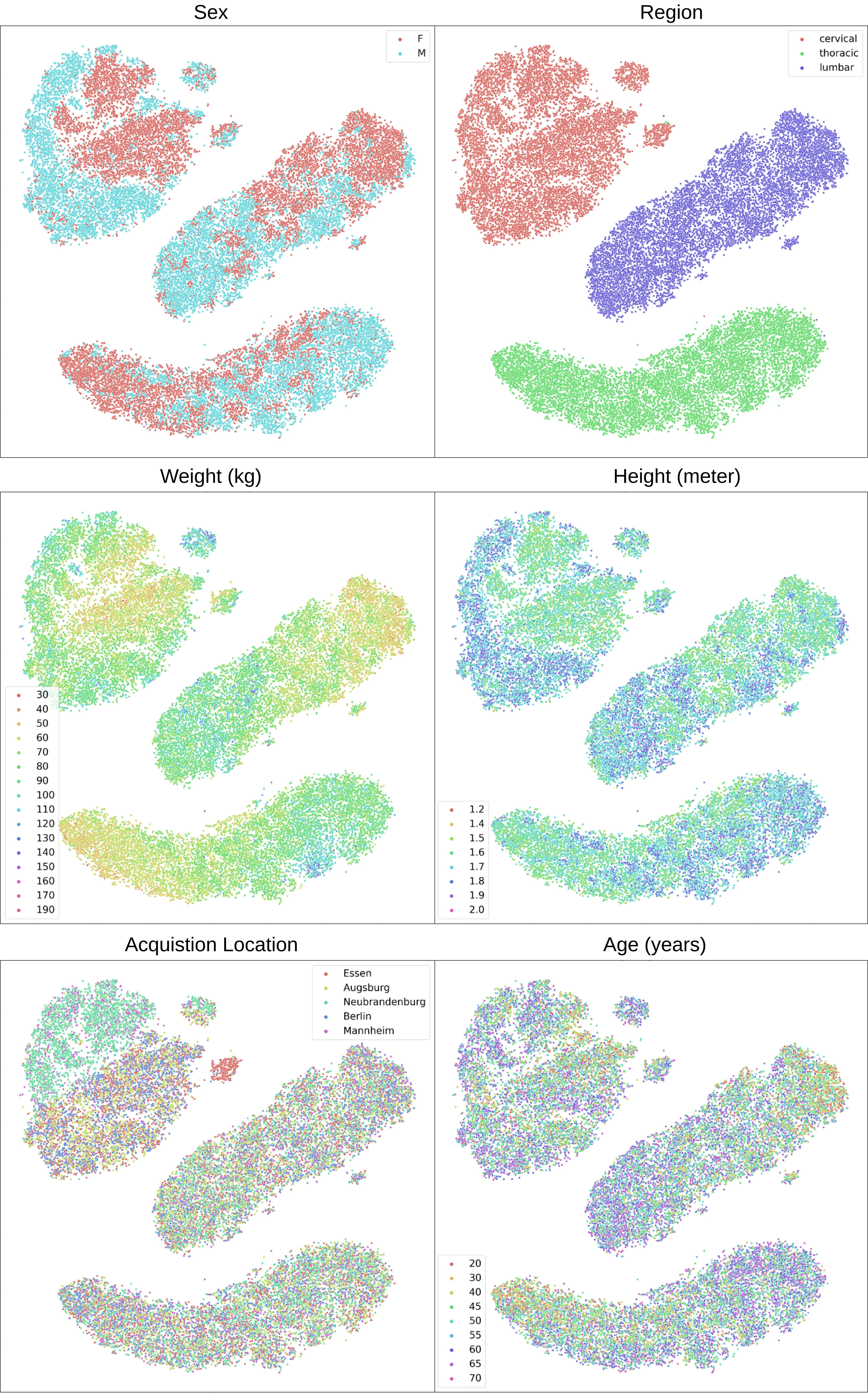}
  \caption{t-SNE plots of StyleGAN embeddings. The embeddings are colored with the patient sex, height, weight, and age. The Region label describes the body region; the acquisition location is the city where the image was recorded. }
\end{figure}
\end{document}